\documentclass[letterpaper, 10 pt, conference]{ieeeconf}  
\IEEEoverridecommandlockouts
\usepackage{cite}
\usepackage{amsmath,amssymb,amsfonts}
\usepackage{graphicx}
\usepackage{textcomp}
\usepackage{xcolor}
\def\BibTeX{{\rm B\kern-.05em{\sc i\kern-.025em b}\kern-.08em
    T\kern-.1667em\lower.7ex\hbox{E}\kern-.125emX}}

\usepackage{float}
\usepackage{array}
\usepackage[caption=false,font=normalsize,labelfont=sf,textfont=sf]{subfig}
\usepackage{stfloats}
\usepackage{url}
\usepackage{verbatim}
\usepackage[normalem]{ulem}
\usepackage{multirow}
\usepackage{soul}
\usepackage{algpseudocode}
\usepackage{algorithm2e}
\usepackage{graphics}
\usepackage{adjustbox}
\usepackage{hyperref}

\usepackage{cuted} 
\usepackage{tabularray}
\usepackage{caption}

\usepackage{booktabs}
\def\BibTeX{{\rm B\kern-.05em{\sc i\kern-.025em b}\kern-.08em
    T\kern-.1667em\lower.7ex\hbox{E}\kern-.125emX}}
\usepackage{balance}
\usepackage{cite}
\usepackage{lipsum, babel}
\begin{document}
\bibliographystyle{IEEEtran}
\newcommand{\RNum}[1]{\lowecase\expandafter{\romannumeral #1\relax}}
\newcommand{\mybluecolor}[1]{{\color{blue}#1}}
\newcommand{\myredcolor}[1]{{\color{red}#1}}

\newcommand{\ASh}[1]{\textcolor{red}{(ASh: #1)}}
\newcommand{\sam}[1]{\textcolor{blue}{(sam: #1)}}

\title{SAFE: Saliency-Aware Counterfactual Explanations for DNN-based Automated Driving Systems}
\author{Amir Samadi, Amir Shirian, Konstantinos Koufos, Kurt Debattista and Mehrdad Dianati
\thanks{The authors are with the Warwick Manufacturing Group,
The University of Warwick, Coventry CV4 7AL. (e-mail:amir.samadi, amir.shirian, konstantinos.koufos, k.debattista and m.dianati@warwick.ac.uk). This research is sponsored by Centre for Doctoral Training to Advance the Deployment of Future Mobility Technologies
(CDT FMT) at the University of Warwick.}}

\maketitle
\thispagestyle{plain}
\pagestyle{plain}
\begin{abstract} 

The explainability of Deep Neural Networks~(DNNs) has recently gained significant importance especially in safety-critical applications such as automated/autonomous vehicles, a.k.a. automated driving systems. CounterFactual~(CF) explanations have emerged as a promising approach for interpreting the behaviour of black-box DNNs. 
A CF explainer identifies the minimum modifications in the input that would alter the model's output to its complement. In other words, a CF explainer computes the minimum modifications required to cross the model's decision boundary. Current deep generative CF models often work with user-selected features rather than focusing on the discriminative features of the black-box model. Consequently, such CF examples may not necessarily lie near the decision boundary, thereby contradicting the definition of CFs. To address this issue, we propose in this paper a novel approach that leverages saliency maps to generate more informative CF explanations. 
Our approach guides a Generative Adversarial Network based on the most influential features of the input of the black-box model to produce CFs near the decision boundary. We evaluate the performance of this approach using a real-world dataset of driving scenes, BDD100k, and demonstrate its superiority over several baseline methods in terms of well-known CF metrics, including proximity, sparsity and validity. Our work contributes to the ongoing efforts to improve the interpretability of DNNs and provides a promising direction for generating more accurate and informative CF explanations\footnote{Source codes are available at: \url{https://github.com/Amir-Samadi//Saliency_Aware_CF}}.
\end{abstract}


\section{Introduction}
\label{sec:intro}

Deep Neural Networks (DNNs) have achieved remarkable success in solving complex tasks, ranging from image recognition \cite{zou2023object} to natural language processing \cite{hirschberg2015advances}. However, their black-box nature has impeded their widespread adoption in safety-critical applications, such as healthcare \cite{mahmud2018applications} and automated driving systems \cite{tampuu2020survey}, where interpretability and transparency are paramount. The advent of Interpretable Artificial Intelligence (IAI) has offered a potential solution to these challenges, and CounterFactual (CF) explanations have gained prominence as a promising IAI approach for revealing in human-understandable terms the underlying rationale behind the decisions of black-box AI systems. A CF example identifies the minimum required changes to the model's input that would alter the model output to its complement. 
\begin{figure}
    \centering
    \includegraphics[width=1.\linewidth]{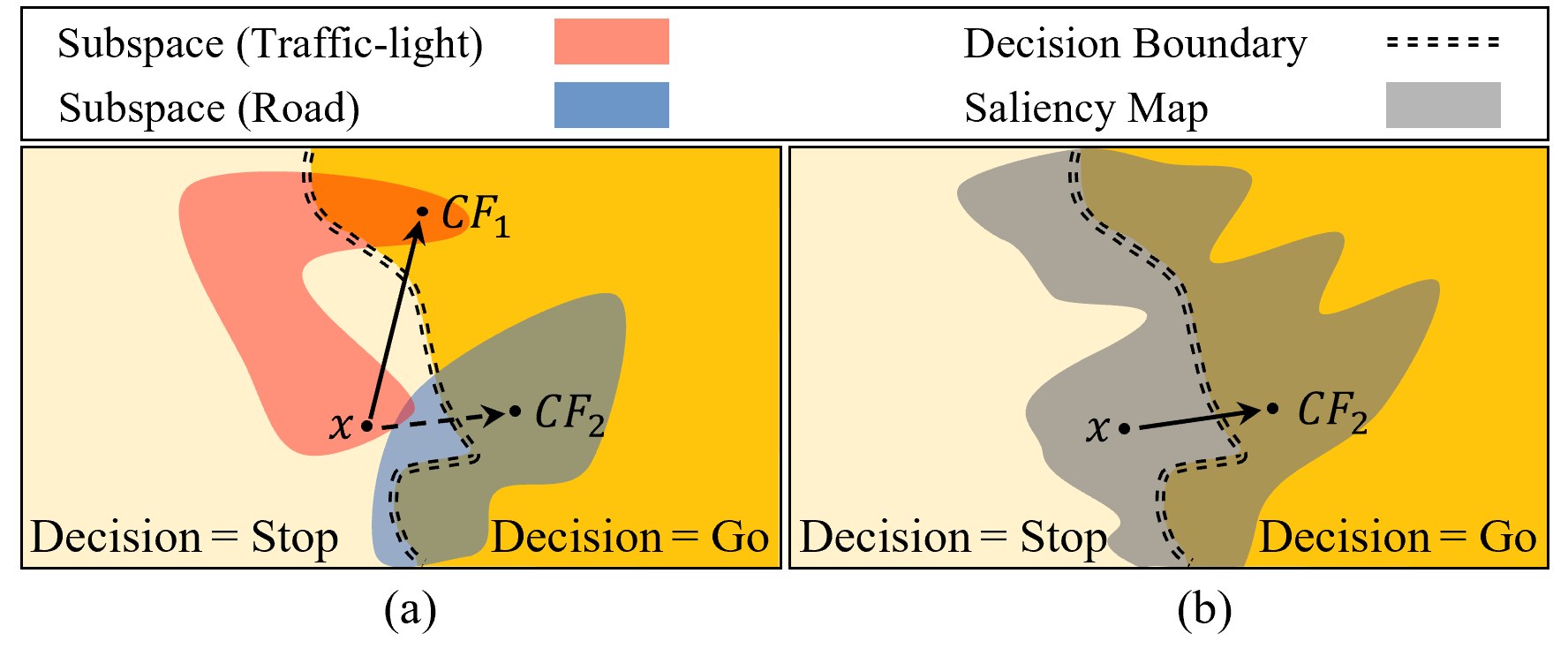}
    \caption{SAFE explores a wider range of possible subspaces compared to previous methods. (a) Previous approaches are limited to predefined semantic subspaces to find the CF explanation, resulting in $\text{CF}_{1}$. (b) While considering a wider subspace, SAFE finds the closest CF  to the decision boundary based on the saliency maps coming from the black-box model, denoted as $\text{CF}_{2}$.
    }
    \label{fig:teaser}
\end{figure}

In recent years, a number of approaches have been proposed for generating CF explanations. Perturbation-based techniques modify the input data by adding or removing specific features~\cite{ivanovs2021perturbation}, while gradient-based methods use the model gradients to identify and alter the most influential features of the input data~\cite{wachter2017counterfactual}. Unfortunately, despite their promising performance with low-dimensional tabular data, these methods suffer from various limitations. On the one hand, perturbation-based methods can be computationally expensive and often generate non-realistic or irrelevant CF instances~\cite{yang2021generative, samadi2023counterfactual}. On the other hand, gradient-based techniques may not be suitable for DNNs due to their non-linearity, particularly when processing high-dimensional image input data, resulting in adversarial examples~\cite{wachter2017counterfactual, goodfellow2014explaining, moosavi2016deepfool}. 

To overcome the above shortcomings, deep generative models, such as Generative Adversarial Networks (GANs)~\cite{goodfellow2020generative}, Variational Auto Encoders (VAEs)~\cite{kim2021counterfactual} and Diffusion Models \cite{jeanneret2022diffusion} have emerged in the literature as promising solutions. Generative models can learn to map the distribution of (high-dimensional) input images between different domains. Transferring a query image from a source domain to another (target) domain is essentially the objective of the CF explanation. Therefore, deep generative models can produce diverse and realistic CF examples for high-dimensional input data~\cite{chou2022counterfactuals}. However, for complex domains, e.g., driving scenarios, generative models tend to generate implausible or semantically invalid examples~\cite{jacob2022steex}. One way to remedy this issue is to limit the potential solutions by encoding the query images into a disentangled latent state and applying their sparse modification~\cite{jacob2022steex, zemni2022octet, rodriguez2021beyond}. This approach involves the manual selection of the input features, which may lead to sub-optimal and biased explanations nonetheless. For example, the Steex method developed in~\cite{jacob2022steex} generates CF examples by modifying groups of pixels specified in the segmented semantic layer. Although this can provide informative explanations, it actually deviates from the definition of CFs; recall that a CF seeks minimal changes in the input. An alternative solution may exist within another semantic feature class that is closer to the decision boundary of the model, see Fig. \ref{fig:teaser}(a) for an example illustration. 
This research gap in generating relevant and unbiased CF explanations near the decision boundary of the model serves as the key motivation for the present study.

To this end, instead of altering user-defined input features, we propose a novel approach that leverages saliency maps to explicitly guide the generative model to focus on the most influential features of the input and generate more effective and unbiased CF explanations. Saliency maps constitute a significant development in advancing the realm of IAI, as they provide insights into the regions of the input that receive the most attention from the black-box model during its decision-making process~\cite{selvaraju2017grad,wang2020score}. That can enhance the quality of the generated CF explanations and achieve a more accurate estimation of the decision boundary, which is crucial for improving the interpretability of DNN models. 
The connection between the saliency map and the decision boundary has been explored in previous studies~\cite{etmann2019connection,mangla2021saliency} and an intuitive illustration of that can be found in Fig.~\ref{fig:teaser}(b). 

The main contribution of the present article lies in integrating saliency maps, extracted from pre-trained off-the-shelf models, into the GAN training process to better understand the discriminative features in the input space. 
The proposed deep generative CF explainer is hereafter referred to as Saliency-Aware counterFactual Explainer (SAFE) and the associated deep generative model as SAcleGAN. The key contributions of this article to the state-of-art are summarised next:

\begin{itemize}
\item Devising a framework for improving the interpretability of DNNs in complex real-world applications by introducing the use of saliency maps encompassing decision boundary information.
\item Introduction of a novel term in the loss function that ensures the proper fusion of saliency information in the SAcleGAN.
\item A comprehensive performance analysis of the SAcleGAN using a complex driving scene dataset, i.e., the BDD. The evaluation demonstrates the superiority of SAcleGAN over several baseline methods in terms of validity, sparsity, and proximity of generated CF examples.
\item A qualitative analysis of the results to gain insights into the strengths and limitations of SAFE. The results provide strong empirical evidence of the effectiveness and reliability of our approach.
\end{itemize} 

The rest of this paper is organised as follows. In Section~\ref{section:related}, we briefly review the related work on IAI and generating CF  explanations. In Section~\ref{section:method}, we describe how to leverage saliency maps to guide the generation of CF explanations. 
Section~\ref{section:results} presents the evaluation results and discusses their implications. Finally, we conclude this work in Section~\ref{section: conlusion}.

\section{Background and Related work}
\label{section:related}
This section provides an overview of the related work in the area of interpretability with a focus on CF explanations. More specifically, we review generative CF methods as they have shown to be more effective on real-world data.

\subsection{Interpretability in Machine Learning}
Interpretability has been a topic of growing interest in the machine-learning community over the past few years. Various methods can increase the interpretability of models, ranging from simple rule-based models and decision trees to more complex techniques such as sensitivity analysis~\cite{samek2016evaluating} and deep learning models~\cite{kim2019grounding}.
These methods become particularly important in safety-critical situations such as autonomous driving applications, where interpretability is essential for reliability, regulatory compliance, and end-user trust~\cite{wachter2017counterfactual}.

Interpretable methods include both post-hoc and intrinsic models. 
Intrinsic methods refer to the models that involve designing and training DNN models with built-in interpretability and post-hoc methods employ auxiliary models to extract explanations from already trained models.
Post-hoc explanations can be global or local, depending on the level of detail they provide. The former provides a holistic view of the main factors driving the decision-making process of the model, and the latter targets to understand the model's behaviour on a specific input~\cite{ribeiro2016should}. 
Various methods, such as saliency maps, LIME~\cite{ribeiro2018anchors}, SHAP~\cite{lundberg2017unified}, partial dependence plots~\cite{ziegel2003elements}, and feature importance rankings~\cite{breiman2001random}, facilitate the generation of both local and global explanations, allowing users to gain insights into model decisions, identify biases or errors, and understand the model's behaviour on individual instances as well as across the entire dataset.

Despite the recent progresses in IAI research, there is still a strong need for better models. The CF explanations are a subset approach to post-hoc explanations and are gaining momentum due to their ability to provide human-understandable insights into the model's behaviour~\cite{wachter2017counterfactual}. Recent works have focused not only on   explaining the model decisions but also on generating CF examples, which are minimally modified versions of the input data that would change the model's output~\cite{wachter2017counterfactual}. 
In the context of automated driving systems (ADS), CF explanations can improve our understanding of why an automated driving model chose a particular action and what features of the input were the most crucial during decision-making~\cite{kim2017interpretable}. For example, CF explanations can indicate whether the colour of a traffic light or the orientation of a pedestrian's body influenced the model's decision to stop or proceed.
Conventional CF explanation techniques employ optimisation methods, such as gradient-based~\cite{wachter2017counterfactual}, genetic~\cite{guidotti2018local, sharma2019certifai, dandl2020multi} and game-theory\cite{ramon2020comparison, rathi2019generating} approaches and perform well with simple tabular data~\cite{wachter2017counterfactual}. However, they struggle to generate visually appealing outputs when dealing with  high-dimensional image data. To address this issue, deep generative CF models have been introduced. 

\subsection{Deep Generative Counterfactual Models}
Deep generative CF explanations are a promising approach to increasing the interpretability of complex decision-making systems, such as DNNs~\cite{AugustinBC022,jacob2022steex,khorram2022cycle}. 
For instance, in an initial work\cite{liu2019generative}, a trained classifier named as AttGAN provides labels for a cycleGAN~\cite{he2019attgan}. This deep generative model then transfers the input image to the target domain serving as a CF explanation example. In a follow-up work~\cite{huber2023ganterfactual}, a starGAN~\cite{choi2018stargan} is employed to generate CF examples for a  trained reinforcement learning-based model. The authors in~\cite{kothiyalutilization} conducted a comparison of different cycleGAN models for generating CF examples focusing on image quality metrics. The majority of works can explain classifiers working on simple images, such as face portraits or featuring a single and centered object.
Nevertheless, generating CF explanations in complex environments, such as those encountered in ADS, can be challenging due to the high diversity and scattered distribution of images in the available driving datasets~\cite{jacob2022steex}. To this end, several works tend to limit the solution space for guiding the learning of generative models.



One way to restrict the solution space utilises an encoder-decoder pair in deep generative models, which effectively maps the input data distribution into a disentangled latent space~\cite{jacob2022steex, zemni2022octet}. This learned feature space ensures that each dimension captures a specific attribute of the data independently of other dimensions, enabling control and understanding of the underlying features of the data that contribute to the generation of the CF example. Deep generative methods can provide minimal and meaningful CF examples by applying optimised manipulations into the latent space. 
However, such CF explanations are obtained from a user-selected feature space rather than being optimised within the entire input feature space. For instance, the STEEX model in~\cite{jacob2022steex} generates CF examples within the driving environment from a  semantic layer selected by the end-user, such as traffic light segmentation, which might end up being suboptimal as the feature space is restricted. Unlike previous methods, in this study, we use saliency map features which lead the generative model to search near the black-box model's decision boundary, resulting in a global minimum solution. 

In this study, avoiding intermediate transferring (to the disentangled latent state) enables our approach to extract more comprehensive features (unknown latent state) and decode them to the queried domain. Meanwhile, for limiting the potential solutions, we use a novel loss function that forces the deep generative model to apply changes merely in salient pixels. Considering that saliency maps encompass decision boundaries~\cite{etmann2019connection, mangla2021saliency}, we have in practice limited the SAcleGAN to apply modifications near the decision boundary regions.

\section{Saliency-Aware counterFctual Explainer}
\label{section:method}

This section provides insights into the SAFE approach that leverages saliency maps to generate CFs using an unpaired dataset in the SAcleGAN. We formulate the problem of generating CFs for a given grey-box model in Section~\ref{subsec: problem_def}, present an overview of the proposed approach in Section~\ref{subsec: SAFE_overview} and finally, describe the SAFE training phase in Section~\ref{subsec: learning}. 

\subsection{Problem Definition}
\label{subsec: problem_def}
Let $x \in \mathcal{R}^{H \times W \times C}$ be an input instance with $H, W$ and $C$ being the height, width and  channel. Also, let $M(\cdot)$ be our grey-box model (having access to some internal hyperparameters), classifying $x$ as $y \in \mathcal{Z}^d$, where $d$ is the number of classes in the grey-box model. Given a target class $y' \neq 
 y $, we shall generate a CF $x' \in \mathcal{R}^{H \times W \times C}$ such that $M(x')=y'$ and $x'$ is as similar as possible to $x$. To this end, the problem in this paper is formulated as leveraging saliency maps to guide the generation of $x'$, i.e., the changes made to $x$ in the search of $x'$ are concentrated in the most salient regions yielding minimal and effective substitutions.

\subsection{Overview of SAFE}
\label{subsec: SAFE_overview}
\begin{figure*}
\vspace{0.15mm}
\centering
\includegraphics[width=0.9\linewidth]{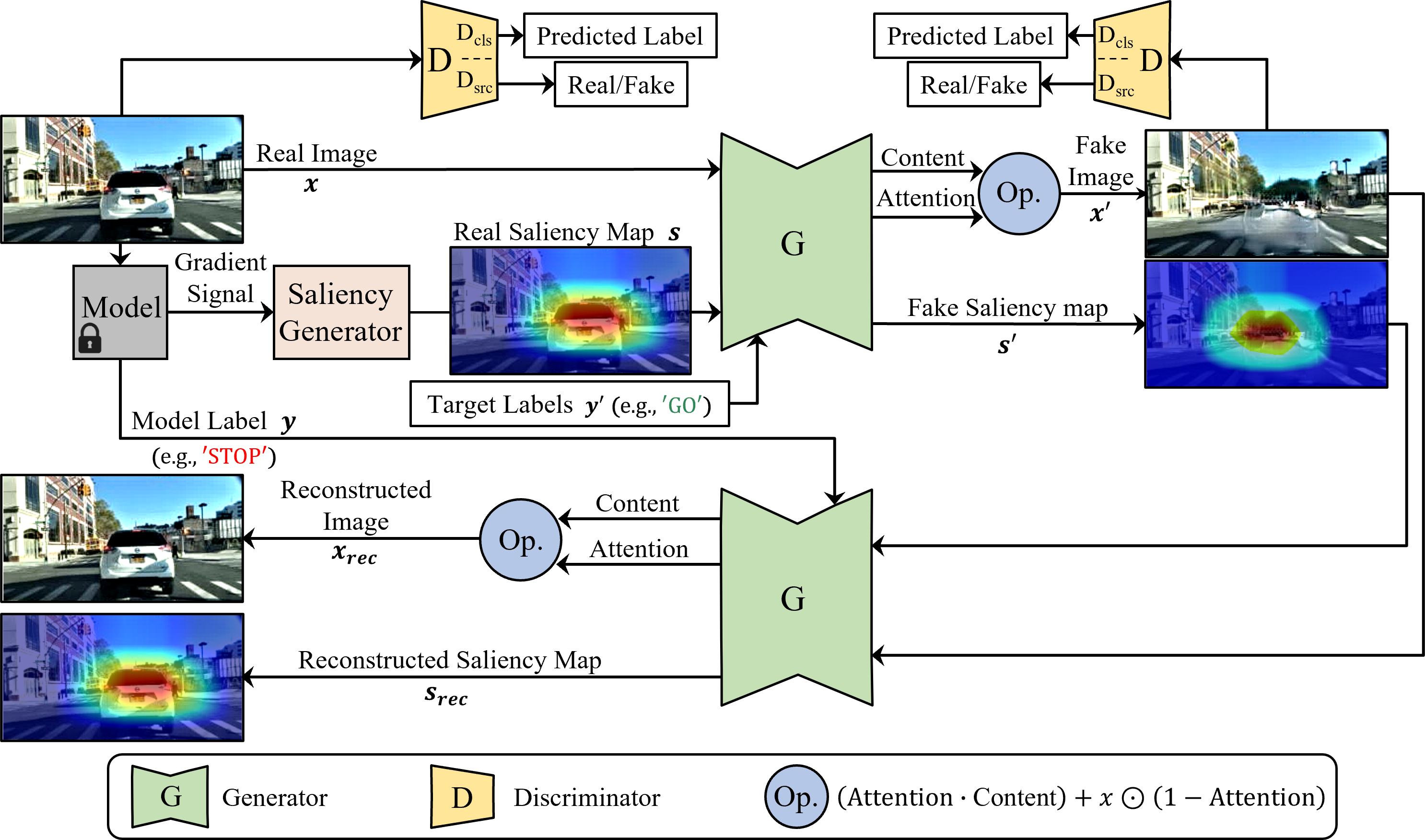}
\caption{Block diagram of the SAcleGAN model.}
\label{fig: method0}
\end{figure*}

In this section, we first discuss the construction of the saliency map, and then detail the procedure of CF generation using the SAcleGAN model, see Fig.~\ref{fig: method0}. Previous studies utilised different GAN-based models to create CFs~\cite{liu2019generative, huber2023ganterfactual,kothiyalutilization,jacob2022steex, zemni2022octet}, but they proved inadequate in uncovering those  associated with the minimal and effective changes. To address this, we use  AttentionGAN~\cite{tang2019attention} as backbone model and introduce a novel loss function term to ensure that the applied changes are minimal within an effective area. Following the conventional GAN architecture~\cite{GoodfellowPMXWOCB14}, AttentionGAN is composed of two competing modules, i.e., the generator $G(\cdot)$ and the discriminator $D(\cdot)$,  iteratively trained to compete against each other in the manner of two-player minimax games. 

\subsubsection{Saliency Map Generation}

 Saliency maps are visualisations used to understand the behaviour of deep learning models in image classification tasks by highlighting the parts of an image used to determine the output class. Recently, gradient-based approaches have been used to compute saliency maps by measuring the impact of each pixel on the output. For an input image $x \in \mathcal{R}^{H \times W \times C}$ classified as $y$, the saliency map generator returns a 2D map $s=S(x,y) \in \mathcal{R}^{H \times W}$ representing the salient parts of the input image with respect to the model's output. In order to generate a saliency map, we employ the Gradient-weighted Class Activation Mapping (Grad-CAM)~\cite{WangWDYZDMH20} method. 

\subsubsection{Generating CF Explanation}
The saliency map along with the original image and the target label enables the generator to extract important features about the grey-box model and alter them to obtain the desired CF. 
Fig.~\ref{fig: method0} illustrates the SAcleGAN training phase, where for a query  image $x$, the grey-box model $M(\cdot)$ provides a label $y$. Accessing the gradient signals of the model (hyperparameters), the Grad-CAM provides the saliency map $s$ of the real image. The Generator network~$G(\cdot)$ then transfers these images into the target domain/label $y'$, i.e., $G(x, s, y') = (x', s')$, where $x'$ and $s'$ denote the CF example (fake image) and its predicted saliency map, respectively. Since paired ground-truth images are not available, the SAcleGAN transfers the provided images back to the initial domain to evaluate the generator's performance, i.e., $G(x', s', y) = (x_{rec}, s_{rec})$ where $x_{rec}, s_{rec}$ are the reconstructed image and saliency map, respectively that are used to train the model. Meanwhile, the discriminator network~$D(\cdot)$ learns to predict image labels, denoted by $D_{cls}(\cdot)$, and distinguish real images from fake ones, showed by $D_{src}(\cdot)$. In Fig.~\ref{fig: method0} one can also find attention and content layers, following the architecture of AttentionGAN in~\cite{tang2021attentiongan}. The attention matrix indicates the pixels that need to be varied in one channel, while the content matrices provide the (target) RGB values of the indicated pixels. The 'Op' component in the figure implements the operation specified in the legend, resulting in the generation of the desired image.



\subsection{Training phase}
\label{subsec: learning}
Let us assume that the real image $x$ is labeled as $y \in Y$ during training, and the generator provides a fake image (CF example) $x'$ that can be classified with the target label $y' \neq y$. The salience maps of the input $x$ and the CF $x'$ are $s$ and $s'$ respectively. The generator's objective is to deceive the discriminator $D$ by generating images that appear to be real. Conversely, the discriminator $D$ aims to enhance its ability to distinguish between generated samples and real data samples. This dynamic interplay encourages both the generator and discriminator to improve their respective performances. Thus, the adversarial loss function is designed in a way to encourage the aforementioned behaviours~\cite{huber2023ganterfactual} :
\begin{equation}
\label{eq: loss_adv}
\begin{gathered}
    \mathcal{L}^D_{{adv}} = \mathbb{E}_{x} \Bigr[D_{src}(x)\Bigr] - \mathbb{E}_{x'} \Bigr[D_{src}(x') \Bigr] \quad  + \\
    \lambda_{gp} \, \mathbb{E}_{\hat{x}}\Bigr[(\lVert \nabla_{\hat{x}} D_{src}(\hat{x}) \rVert_2-1 )^2\Bigr].
\end{gathered}
\end{equation}
\begin{equation}
\label{eq: x_hat}
    \hat{x} = \alpha  x + (1 - \alpha) x', 
\end{equation}
where $\alpha$ is a uniform random variable in $(0, 1)$, $||\cdot||_2$ the Euclidean norm of a vector, $\nabla_{\hat{x}}$ is the gradient of $\hat{x}$ and $\hat{x}$ is a linear combination of $x$ and $x'$. 

While the first two terms in Eq.~\ref{eq: loss_adv} guide the discriminator to distinguish real images from fake ones, the gradient penalty loss term 
\begin{equation}
\mathcal{L}_{gp} = \mathbb{E}_{\hat{x}}\Bigr[(\lVert \nabla_{\hat{x}} D_{src}(\hat{x}) \rVert_2-1 )^2\Bigr]
\end{equation}
prevents the discriminator from being trapped near trivial solutions~\cite{gulrajani2017improved}. It  ensures that the discriminator's gradients are bounded for better convergence calculated by choosing a random point $\alpha$ along the direction connecting a real image sample and a generated one, see Eq.~\ref{eq: x_hat}, and then taking the  gradient w.r.t. that point. Intuitively, minimising the loss $\mathcal{L}_{gp}$ for a random point between real and fake images, ensures that the discriminator has smooth gradients along the data manifold, preventing it from ignoring or focusing too much on individual samples and improving the overall stability of the GAN training process.

To evaluate whether the CF examples can be classified as $y'$ by the grey-box model $M$, the discriminator should learn the mapping of images to corresponding labels. This can be achieved through the following loss function:
\begin{equation}
\begin{gathered}
    \mathcal{L}^y_{{cls}} = \mathbb{E}_{x,y} \Bigr[(-\log D_{cls}(y|x)\Bigr].
\end{gathered}
\end{equation}

A linear combination of the two loss functions is used to train the discriminator network:
\begin{equation}
\label{eq: L_d}
\begin{gathered}
    \mathcal{L}_D = -\mathcal{L}_{adv}^D + \lambda_{{cls}}\mathcal{L}_{{cls}}^y.
\end{gathered}
\end{equation}

To further encourage the generator, the following loss functions are utilised to generate real-like images. To achieve this objective, the generator is penalised if it fails to deceive the discriminator in differentiating real images from generated ones and mis-classifies generated images into the target labels. 
\begin{equation}
\begin{gathered}
    \mathcal{L}^G_{{adv}} = \mathbb{E}_{x'} \Bigr[D_{src}(x) \Bigr].\\
    \mathcal{L}^{y'}_{{cls}} = \mathbb{E}_{x',y'}\Bigr[-\text{log}D_{cls}(y'|x')\Bigr].
\end{gathered}
\end{equation}

Moreover, since the ground truth of CF examples is unavailable, following the SAcleGAN approach, we transfer back the generated CF example (fake image) into the initial domain and compare it with the query image. The following loss penalises the generator model for adding excessive artifacts to enforce forward and backward consistency:
\begin{equation}
\begin{gathered}
\mathcal{L}_{rec} = \mathbb{E}_{x,y,y'}\Bigr[ \lVert{ (x,s)^T -  G(G(x, s,y'),y)^T} \lVert_1 \Bigr], 
\end{gathered}    
\end{equation}
where the transpose operator $(\cdot,\cdot)^T$ turns the two matrices into a column vector. 





To incorporate the saliency maps into the learning phase, we propose a novel term, $\mathcal{L_\text{fuse}}$, into the loss function. This  term ensures that the saliency map information is fused into the generator model properly. Penalising the generator for modifying the  non-salient features, $(1-s)$, leads the model to perceive the salient region and alter only that region's pixels towards the generation of the CF example:
\begin{equation}
\label{eq:lfuse}
    \mathcal{L}_{fuse} = \mathbb{E}_{x,y,y'}\Bigr[ || (x-x')\odot (1-s) ||_1\Bigr], 
\end{equation}
where $\odot$ denotes point-wise multiplication of matrices.

Finally, a linear combination of the three introduced loss functions shapes the generator model's objective $\mathcal{L_\text{G}}$:
\begin{equation}
\label{eq: L_G}
    {\mathcal{L}_{G}} = -\mathcal{L}^G_{adv} + \lambda_{{cls}} \mathcal{L}_{{cls}}^{y'} + \lambda_{{gp}} \mathcal{L}_{{gp}} + \lambda_{{fuse}} \mathcal{L}_{{fuse}}.
\end{equation}

The pseudocode under Algorithm~\ref{alg: alg} summarises the SAFE approach  generating the CF examples.

\RestyleAlgo{ruled}

\SetKwComment{Comment}{\textcolor{teal}{/* }}{ \textcolor{teal}{/* }}
\SetKwInput{kwReturn}{return}
\begin{algorithm}[hbt!]
\caption{Pseudo-code of SAFE 
}\label{alg: alg}
\Comment{~~~~~~~~\textcolor{teal}{Training Procedure}~~~}
\KwData{Input image $x$ labeled as $y$, target label $y'$}
\KwResult{CF instance $x'$}
$N \gets 0$\\ 
\For{$x$ in data}{
    ${l}_{class} \gets \mathcal{L}_{cls}(D_{cls}(x), M(x))$;\\
    ${l}_{real} \gets \mathcal{L}^D_{adv}(D_{src}(x))$;\\
    $s \gets \text{Grad-cam}(x, y, y')$;\\
    $x', s' \gets G(x, s, y')$;\\
    ${l}_{fake} \gets \mathcal{L}^D_{adv}(D_{src}(x'))$;\\
    $\alpha \gets rand(0,1)$;\\  
    $\hat{x} \gets (\alpha  x + (1 - \alpha)  x')$;\\
    $l_{gp} \gets \mathcal{L}_{gp}(D_{src}(\hat{x}))$;\\
    $l_{D} \gets  -l_{real} + l_{fake} + \lambda_{cls}  l_{class} + \lambda_{gp}  l_{gp}$;\\
    $D \gets ADAM(D, l_{D}, \mathit{learning\mbox{-}rate});$\\
    $N \gets N+1$;\\
    \If{$N~\text{is}~5$}{
        $x', s' \gets G(x, s, y')$;\\
        $l_{fake} \gets D_{src}(x')$;\\
        ${l}_{class} \gets \mathcal{L}_{class}(D_{cls}(x'), M(x'))$;\\
        $x_{rec}, s_{rec} \gets G(x', s', M(x))$;\\  
        $l_{rec} \gets \mathcal{L}_{rec}(x,s,x_{rec}, s_{rec})$;\\
        $l_{fuse} \gets \mathcal{L}_{fuse}(x, x', s)$;\\
        $l_{G} \gets -l_{fake} + \lambda_{cls} l_{class} + \lambda_{gp} l_{gp} + \lambda_{fuse} l_{fuse}$;\\      
        $G \gets ADAM(G,l_{G}, \mathit{learning\mbox{-}rate});$\\
        $N \gets 0$\\ 
    }
}
\end{algorithm}

\section{Evaluations}
\label{section:results}

In this section, we carry out an analysis of the effectiveness of the SAFE model on the BDD dataset, and compare its performance against  state-of-the-art methods~\cite{jacob2022steex, zemni2022octet,liu2019generative,huber2023ganterfactual,kothiyalutilization}. The implementation details of the SAcleGAN are presented in Section~\ref{sec:ID}. We employ various metrics to measure the quality of the generated CF explanations, which are presented in Section~\ref{sec:  metric}. After that we proceed with the description of the selected dataset in Section~\ref{sec: BDD} and the discussion of the performance evaluation results in Section~\ref{sec:PE}. Finally, we conduct in Section~\ref{sec: Quality} a qualitative study that illustrates how well the generated CFs convey explanations comprehensible to humans.

\subsection{Implementation Details}
\label{sec:ID}
All input images are resized to $128 \times 256$ pixel resolution. To ensure the robustness of our findings, we repeat our experiments five times with different seeds and report the mean values of the performance metrics. For all network models we use the Adam optimiser with a learning rate of $10^{-4}$. The network hyper-parameters and loss coefficients are explored heuristically and set to $\lambda_{cls}=1, \lambda_{gp}=10, \lambda_{rec}=10, \lambda_{fuse}=1~\text{and}~\lambda_{fuse}=5$ (see Eq. \ref{eq: L_d} and \ref{eq: L_G}) for all experiments. The SAcleGAN employs the limited modified generator network and the pure discriminator network of attentionGAN~\cite{tang2019attention} models, and for the subject grey-box model we use ResNet50, see Fig.~\ref{fig: method0}. Furthermore, The 'Saliency Generator' component of our approach utilises the Grad-CAM method~\cite{selvaraju2017grad}. To execute our implementation, we employ Pytorch framework on an NVIDIA RTX-3090 GPU.

\subsection{Performance Metrics}
\label{sec: metric}

In order to assess the quality of the generated CF examples, we measure their adherence to well-known CF criteria namely  \textbf{Proximity}, \textbf{Sparsity}, and \textbf{Validity}. Proximity refers to the similarity or closeness between the query image and its corresponding CF instance, which can be calculated by measuring the mean value of pixels that are modified. Sparsity refers to the extent to which the changes to the generated CFs are minimal and focus only on a small subset of features. 
Validity measures the success rate of the method generating CFs as being equal to the percentage of generated CF examples that altered the model's output to the target label.

Apart from the traditional performance metrics, one can find in the literature  {\it{generative}} type of metrics that assess the visual realism of the generated CF explanations, such as the \textbf{FID}, \textbf{LPIPS}, \textbf{KID}, and \textbf{IS}. FID measures the similarity between generated and real images based on the statistical properties of their feature representations. LPIPS quantifies perceptual differences between images using high-level visual features. KID measures the dissimilarity between feature distributions of generated and real images using the Maximum Mean Discrepancy method, and, IS evaluates the quality and diversity of generated images by comparing them to the real-world distribution of images.



\subsection{Berkeley DeepDrive Dataset (BDD)}
\label{sec: BDD}
The BDD100k dataset consists of 100,000 detailed images depicting diverse driving scenes~\cite{yu2020bdd100k}. 
The grey-box DNN model used in our experiments is trained to predict actions such as "Move Forward" and "Stop/Slow down" on the BDD-OIA dataset~\cite{xu2020explainable}, which consists of 20,000 scenes specifically selected and annotated with high-level actions from the BDD100k dataset. To generate CF examples, both the SAFE and baseline explainer models are fed with the BDD100k dataset.

\subsection{Performance Evaluation}
\label{sec:PE}
Table~\ref{tab: BDD_0} contains the comprehensive performance comparison results between the proposed model, SAFE, and two well-known cycleGAN models, namely starGAN~\cite{choi2018stargan} and attentionGAN~\cite{tang2021attentiongan}. For the SAFE model, we investigate the effectiveness of the newly introduced loss function, $\mathcal{L_\text{fuse}}$, in Eq.~\eqref{eq:lfuse} for two values of the associated coefficient $\lambda_{fuse}=1$ and $\lambda_{fuse}=5$, indicated by SAFE$_1$ and SAFE$_{5}$ respectively. To ensure a fair comparison with previous studies \cite{liu2019generative,huber2023ganterfactual,kothiyalutilization}, the baseline models are fed with the decision model's output instead of the ground-truth labels. This approach allows for improvements in the proximity, sparsity, and validity metrics of the baseline models. By examining the Table~\ref{tab: BDD_0}, we observe that SAFE outperforms the baselines in terms of validity  substantially, while exhibiting negligible differences in the generative metrics that assess the realism of the generated images (FID, LPIPS, KID, IS). Notably, the effect of the term $\mathcal{L_\text{fuse}}$ in the loss function is evident, where for the lower value of $\lambda_{fuse}$,  SAFE applies more modifications to the query image, resulting in higher validity and greater distance from the initial image. Conversely, a higher value of $\lambda_{fuse}$ leads to a different trade-off where modifications are reduced, resulting in lower validity and a closer resemblance to the initial image.

\begin{table}
\centering
\vspace{1mm}
\caption{Performance comparison of SAFE with cycleGAN models. The direction of arrows preceding the metric indicates which values are desirable (low $\downarrow$, or high $\uparrow$). The best model is highlighted in bold, while the second-best model is underlined. Proximity, sparsity, and validity are denoted by prx, sprs, and vld, respectively. Note that \textit{CF-} represents the counterfactual variant of the model.}
\label{tab: BDD_0}
\begin{tblr}{
  row{1} = {m}{l},
  column{1} = {l}{2.11cm},
  column{2-8} = {c}{.45cm},
  hline{1-2,6} = {-}{},
}
Method           & $\downarrow$FID  & $\downarrow$LPIPS & $\downarrow$KID & $\uparrow$IS   &$\downarrow$Prx & $\downarrow$Sprs & $\uparrow$Vld \\
CF-StarGAN       & \textbf{19.31}  & \textbf{0.117}    & 0.067           & 3.151           & 0.039             & 0.999         & 0.564 \\
CF-AttentionGAN        & \uline{28.90}   & \uline{0.196}     & 0.072           & \uline{3.120}   & \textbf{0.009}    & 0.574         & 0.551    \\
$\text{SAFE}_1$  & 28.45           & 0.204             & \uline{0.064}   & 3.081           & \uline{0.021}     & \uline{0.413} & \textbf{0.938}    \\
$\text{SAFE}_{5}$ & 29.21           & 0.225             & \textbf{0.059}  & \textbf{3.242}  & 0.026             & \textbf{0.409}& \uline{0.914}      
\end{tblr}
\end{table}


\begin{table}[t!]
\centering
\caption{Performance comparison of SAFE with CF explainers. The direction of the arrow preceding the metric indicates which values are desirable (low $\downarrow$, or high $\uparrow$). The best performing model is indicated in bold. The validity metric is denoted by vld.}
\label{tab: BDD_1}
\begin{tblr}{
  column{even} = {c},
  column{3} = {c},
  hline{1-2,5} = {-}{},
}
Method          & $\downarrow$ FID & $\downarrow$LPIPS & $\uparrow$Vld  \\
STEEX           & 61.35           & 0.451             & \textbf{0.973} \\
OCTET           & 54.95           & 0.422             & 0.961          \\
$\text{SAFE}_1$ & \textbf{28.45}  & \textbf{0.204}    & 0.938          
\end{tblr}
\end{table}

\begin{figure}[b!]
\centering
\vspace{-6mm}
\includegraphics[width=\linewidth]{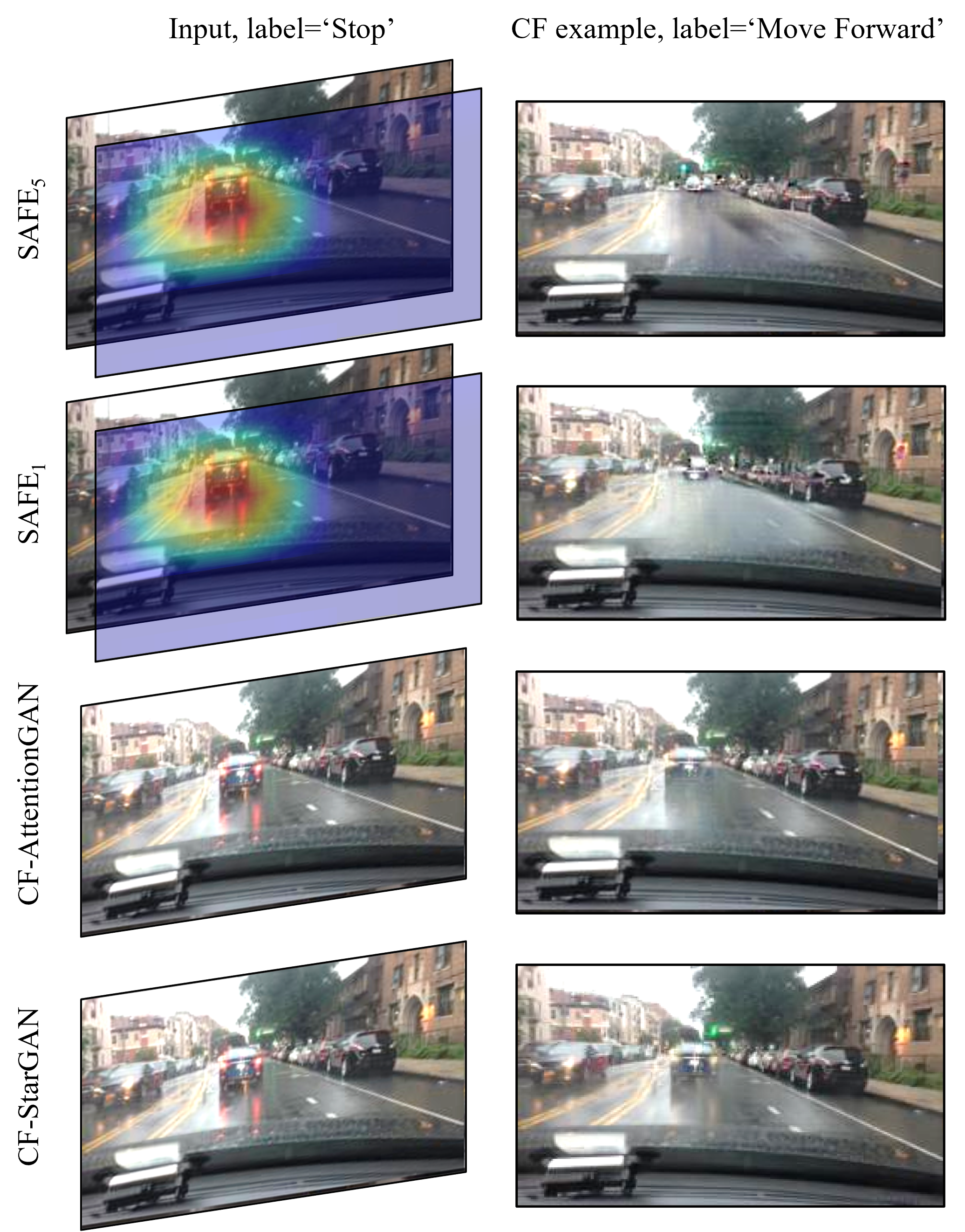}
\caption{Performance comparison by visual inspection between the SAFE and two cycleGAN baseline models.}
\label{fig: BDD}
\end{figure}


\begin{figure*}
\centering
\includegraphics[width=\linewidth]{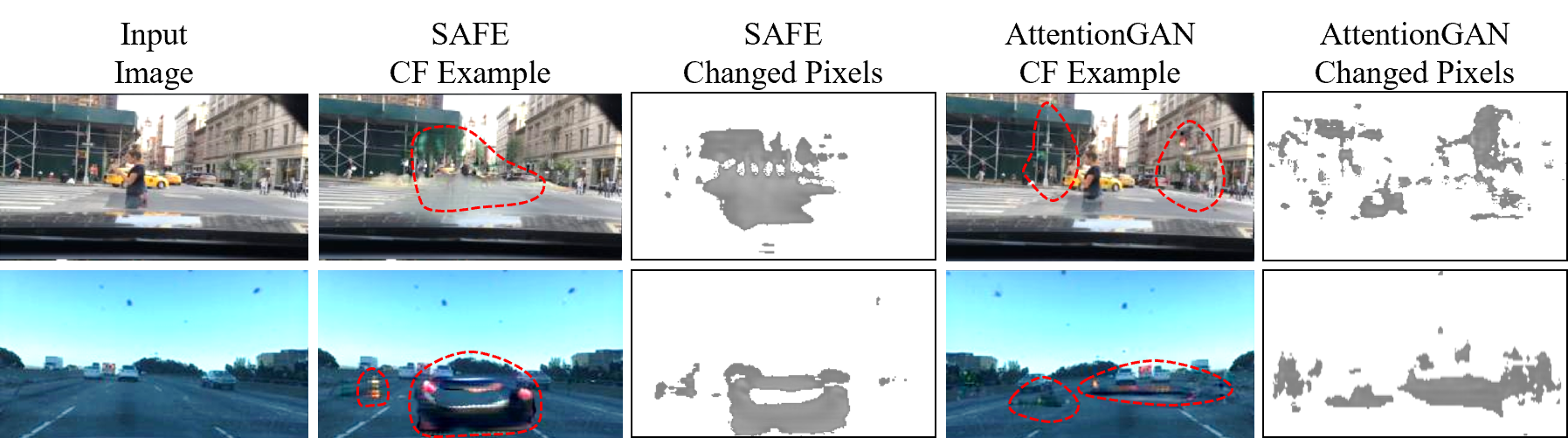}
\caption{Performance comparison by visual inspection between the SAFE and the AttentionGAN baseline model. The target label is 'GO' for the top row and 'STOP' for the bottom row.}
\label{fig: quality}
\end{figure*}
\subsection{Quality of Counterfactual Explanations}
\label{sec: Quality}

Table \ref{tab: BDD_1} presents the performance comparison between SAFE and  the latest state-of-the-art CF explainers, namely STEEX \cite{jacob2022steex} and OCTET \cite{zemni2022octet} based on the results reported in their respective papers. Upon examining this table, we do not observe any significant difference in terms of the explainer model's validity performance, however, when it comes to the realism metrics, our approach demonstrates notable improvements. Specifically, it achieves a 53~\% improvement in LPIPS and a 55~\% improvement in FID metrics as compared to the state-of-the-art CF explainers STEEX and OCTET. These results highlight the superior quality and realism of the CF examples generated by the SAFE model. It is worth noting that both baseline methods, STEEX and OCTET, modify disentangled latent states using user-selected features. This approach may result in a counterfactual example that is far from the decision boundary of the grey-box DNN model, as discussed in sections \ref{sec:intro} and \ref{section:related}. 


Fig.~\ref{fig: BDD} illustrates a visual comparison for the quality of the generated CF examples between SAFE and the baseline cycleGAN models. Given an input labelled as 'Stop' by the decision model (Fig.~\ref{fig: BDD}, first column), the SAFE model leverages a saliency map (depicted as an image overlay) and fades the front vehicle completely to change the decision to 'Go' (Fig.~\ref{fig: BDD}, second column). On the contrary, the baseline models encounter difficulties in achieving such clear modifications in the original image and fail to generate CF examples with similar effectiveness. This visual comparison highlights the superior performance of SAFE in generating meaningful, clear and effective CF explanations as compared to the baseline models.

Fig.~\ref{fig: quality} (top row) illustrates another driving scene showcasing the generation of a CF example that alters the original decision from 'STOP' to 'GO'. The visualisation reveals that SAFE has faded both the pedestrian and the yellow participant vehicle, and moreover, it alters the street sign to a green traffic light. Based on these modifications, we can infer that, under these circumstances, the automated vehicle should proceed forward, taking into account factors such as an obstacle-free urban intersection and the presence of a green traffic light signal. 
The bottom row of Fig.~\ref{fig: quality} presents a different scene in a motorway, wherein the grey-box decision model would only stop if there is a leading vehicle nearby. To summarise, these CF examples effectively highlight the necessary modifications required in the input image to change the automated vehicle's decision, thereby indicating that the DNN model has learned to focus on relevant features during training. This is also evident from the distribution of changed pixels in Fig.~\ref{fig: quality}, where one can see that the SAFE model has successfully concentrated on the important features within the input image, which can be attributed to the utilisation of the saliency maps.
For both driving scenarios we have also generated the CF examples and the changed input pixels by the AttentionGAN~\cite{tang2021attentiongan} model. 
The latter did not succeed in generating plausible CFs as can be also reflected by the scattered distribution of the changed pixels across the image.

\section{Conclusions}
\label{section: conlusion}


This paper designed a novel method named after SAFE for generating counterfactual (CF) explanations leveraging a cycle-GAN guided by saliency maps. 
The latter was adopted to highlight the most salient regions of the input image that play a decisive role in determining the output of the machine learning model and subsequently help alter only those regions given the target (complement) label. The proposed method significantly advanced the state-of-art models in terms of  validity (the higher, the better) and sparsity (the lower, the better) of the generated CF examples using the Berkeley DeepDrive dataset. At the same time, the generated CFs were more interpretable and clear to understand by visual inspection. We believe that this method has the potential to strengthen our trust and facilitate the adoption of machine learning models in real-world applications, such as autonomous vehicles and automated driving systems in general, by providing more transparent and interpretable explanations for their  predictions and decision-making. In the future, it is worth validating the performance of SAFE with other datasets too. 

\bibliography{ref}
\end{document}